\pdfoutput=1

\documentclass[11pt]{article}

\usepackage[final]{acl}

\usepackage{times}
\usepackage{latexsym}

\usepackage[T1]{fontenc}

\usepackage[utf8]{inputenc}

\usepackage{microtype}

\usepackage{inconsolata}

\usepackage{graphicx}
\usepackage{mdframed}
\usepackage{amsfonts}
\usepackage{multirow}
\usepackage{amsmath}
\newcommand{\minisection}[1]{\vspace{0.04in} \noindent {\bf #1}\ \ }

\title{RAG4ITOps: A Supervised Fine-Tunable and Comprehensive RAG Framework for IT Operations and Maintenance}

\author{
 \textbf{Tianyang Zhang\textsuperscript{1}},
 \textbf{Zhuoxuan Jiang\textsuperscript{2}}\thanks{Corresponding author.},
 \textbf{Shengguang Bai\textsuperscript{1}},
 \textbf{Tianrui Zhang\textsuperscript{3}},
 \textbf{Lin Lin\textsuperscript{4}},
\\
 \textbf{Yang Liu\textsuperscript{5}}\and
 \textbf{Jiawei Ren\textsuperscript{1}}
\\
 \textsuperscript{1}Learnable.ai, Shanghai, China\\
 \textsuperscript{2}Shanghai Business School, Shanghai, China\\
 \textsuperscript{3}University of North Carolina Greensboro, Greensboro, NC, USA\\
 \textsuperscript{4}Skema Business School, Paris, France\\
 \textsuperscript{5}North Carolina Central University, Durham, NC, USA\\
 \small{
\url{tzhang@aggies.ncat.edu},
\url{jzx@sbs.edu.cn},
\url{shengguang.bai@learnable.ai}
}
}

\begin{document}
\maketitle

\begin{abstract}
With the ever-increasing demands on Question Answering (QA) systems for IT operations and maintenance, an efficient and supervised fine-tunable framework is necessary to ensure the data security, private deployment and continuous upgrading. Although Large Language Models (LLMs) have notably improved the open-domain QA's performance, how to efficiently handle enterprise-exclusive corpora and build domain-specific QA systems are still less-studied for industrial applications. In this paper, we propose a general and comprehensive framework based on Retrieval Augmented Generation (RAG) and facilitate the whole business process of establishing QA systems for IT operations and maintenance. In accordance with the prevailing RAG method, our proposed framework, named with RAG4ITOps, composes of two major stages: (1) Models Fine-tuning \& Data Vectorization, and (2) Online QA System Process. At the Stage 1, we leverage a contrastive learning method with two negative sampling strategies to fine-tune the embedding model, and design the instruction templates to fine-tune the LLM with a Retrieval Augmented Fine-Tuning method. At the Stage 2, an efficient process of QA system is built for serving. We collect enterprise-exclusive corpora from the domain of cloud computing, and the extensive experiments show that our method achieves superior results than counterparts on two kinds of QA tasks. Our experiment also provide a case for applying the RAG4ITOps to real-world enterprise-level applications.
\end{abstract}

\section{Introduction}
In recent years, the field of IT operations and maintenance has become increasingly significant due to the rapid expansion of massive data and complex IT systems, such as in cloud computing and telecommunications~\citep{liu2023logprompt}. Efficient IT operations and maintenance are critical for providing the high-quality performance, reliability, and security for customers in the business area~\cite{du2017deeplog, guo2023owl}. 
\begin{figure}[t]
\centering
  \includegraphics[width=0.48\textwidth]{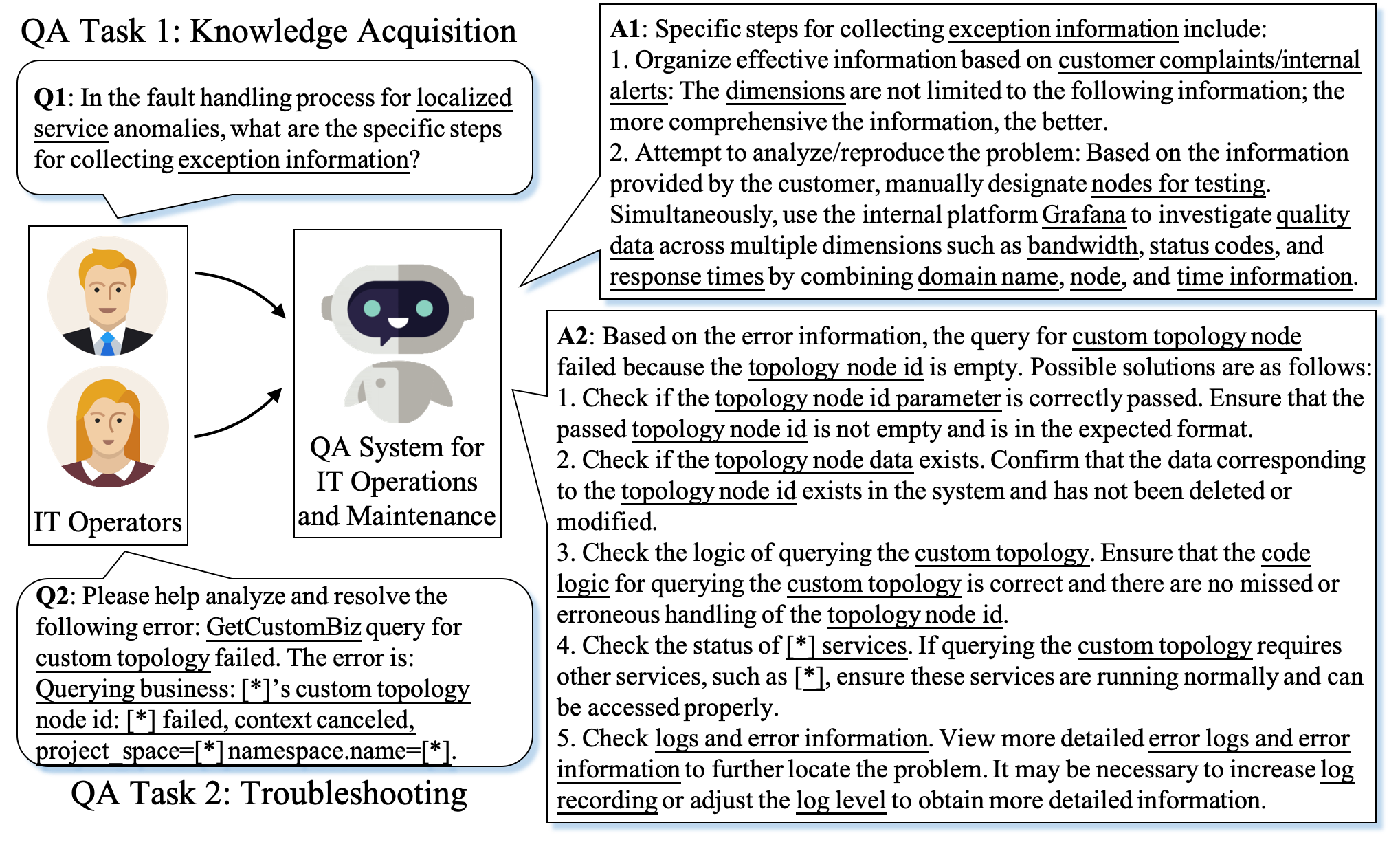}
  \vspace{-4ex}
  \caption{Two examples of typical and important QA scenarios for IT operations and maintenance. The words with underlines are domain-specific terminologies, and the $[\ast]$ represents enterprise-exclusive terms, e.g. status codes or service names.}
  \label{task}
\vspace{-2ex}
\end{figure}

Traditionally, to operate and maintain those systems, it highly depends on IT operators' personal experience, while often leading to difficulties in incident management, problem resolution, and maintaining service quality~\citep{jantti2017proactive}. Later with the advancements of QA techniques, some QA systems are developed, and IT operators can leverage them to retrieve useful information and make a plan on troubleshooting efficiently in a natural-language human-machine interacting manner~\citep{huang2023logqa, jantti2017proactive, galup2009overview}. As shown in Figure~\ref{task}, the two typical and important QA tasks are \emph{Knowledge Acquisition} and \emph{Troubleshooting}~\citep{rijal2022aiops}. The former is usually for junior IT operators to promote their experience, while the latter is for senior ones to obtain guidance on resolving difficult software and hardware faults during their daily work. Therefore, QA systems have become greatly important in contemporary IT operations and maintenance.

To build the QA systems for IT operations and maintenance, we observed numerous examples such as those in Figure~\ref{task}. Some characteristics and challenges can be summarized as follows: First, the QA utterances contain many technical terminologies (e.g., status codes, service names and other underlined words/$[\ast]$ as illustrated in Figure~\ref{task}), and their semantics are exclusive to a specific domain or even an enterprise. Therefore, the enterprise-exclusive semantics should be thoroughly modeled. Second, in terms of data forms, a vast amount of enterprise-exclusive documents, guides and manuals should be processed and modeled into a uniformed format to support the continual upgrading of QA systems. Third, the difficulties of various QA tasks are distinct. For example, QA for knowledge acquisition requires a system only to answer the question with straightforward information, while the QA for troubleshooting demands a much longer answer that involves referring to multiple resources. To this end, all the above challenges lead to a complex problem of how to build an efficient framework that addresses exclusive data and specific QA tasks.

Intuitively, some open-domain QA systems that are trained on massive public corpora can be leveraged to further fine-tune on domain-specific corpora and tasks, especially with the recent breakthroughs of LLMs~\citep{chowdhery2023palm, bai2023qwen, openai2023gpt, brown2020language} such as BERT, LLaMA-3~\citep{touvron2023llama}, Qwen, and ChatGLM3~\citep{zeng2022glm, du2022glm}. However, those LLMs are still too general to be adaptive for distinct QA tasks, or efficiently support the continuous data or/and system upgrading in real-world applications.

To address the above-mentioned problems, in this paper, we leverage the idea of Retrieval Augmented Generation (RAG) which can strengthen LLMs~\citep{gao2023retrieval}, and propose a comprehensive RAG framework specific for the domain of IT operations and maintenance, named with RAG4ITOps. In accordance with
the prevailing RAG methodology, our framework composes of two stages: (1) Models Fine-tuning \& Data Vectorization, and (2) Online QA System Process. The framework features include a data pipeline for efficiently processing multi-source and multi-form enterprise-exclusive corpora, a domain knowledge augmented embedding model for modeling exclusive semantics, and a supervised fine-tuned LLM which can support grounded QA tasks. 

More specifically, to build a QA system by using the proposed RAG4ITOps, firstly the enterprise-exclusive corpora should be collected and preprocessed in advance. After several automatic steps of data cleaning, chunking and distillation, we can obtain a high-quality set of text chunks and two datasets with annotations for fine-tuning the embedding model and the LLM respectively. To better distinguish the QA tasks and model enterprise-exclusive semantics, we fine-tune the embedding model by adopting the contrastive learning~\citep{gao2021simcse} with Homogeneous In-Batch Negative Sampling (HIS)~\citep{zhang2023retrieve} and Auxiliary Hard Negative Sampling (AHNS) strategies. Then the set of text chunks are embedded into vectors by using the fine-tuned embedding model, and stored in the vector database. As to the LLM, we also fine-tune it with QA pairs by adopting a Retrieval Augmented Fine-Tuning method. The details can be found in the Methodology section.

In summary, the proposed RAG4ITOps is supervised fine-tunable on both exclusive data and grounded QA tasks. Note that due to the nature of RAG mechanism, the vector database can be easily updated by inserting new data, instead of frequently refine-tuning the LLM. And the LLM can dynamically incorporate retrieved top-k contents from the database, which are always latest and most relevant. In this way, the requirement of continuous data or/and system upgrading is fulfilled with a low cost. We collect enterprise-exclusive corpora from the domain of cloud computing, and the experiment results show that our framework can achieve superior performance than counterparts on both QA tasks. Our experiment also provides a case of how to apply the RAG4ITOps into real-world enterprise-level applications.

The contributions of this paper include:
\begin{itemize}
\setlength{\itemsep}{0pt}
  \setlength{\parskip}{0pt}
  \setlength{\itemindent}{0em}
    \item To satisfy the ever-increasing demands on QA systems for IT operations and maintenance, we propose a comprehensive RAG-based framework named RAG4ITOps. This framework facilitates the business process of data modeling and model fine-tuning.
    \item The proposed framework composes of two stages:  (1) Models Fine-tuning \& Data Vectorization, and (2) Online QA System Process. We leverage several latest techniques to fine-tune the embedding model and LLM, including the contrastive learning method with Homogeneous In-Batch Negative Sampling and Auxiliary Hard Negative Sampling strategies, the design of instruction templates, and a Retrieval Augmented Fine-Tuning method.
    \item The RAG4ITOps features: (1) a data pipeline that can automatically process multi-source and multi-form enterprise-exclusive corpora, (2) a fine-tuned embedding model for modeling enterprise-exclusive semantics, and (3) a fine-tuned generative LLM which can support distinct QA tasks.
    \item Real-world corpora of IT operations and maintenance for cloud computing were collected, and extensive experiments demonstrate that all the components of RAG4ITOps effectively improve the performance of distinct QA tasks. The experiments also establish a case for our framework to be applied across various enterprise-level applications.
\end{itemize}

\begin{figure*}[t]
\centering
  \includegraphics[width=1\textwidth, height=0.3\textheight, keepaspectratio]{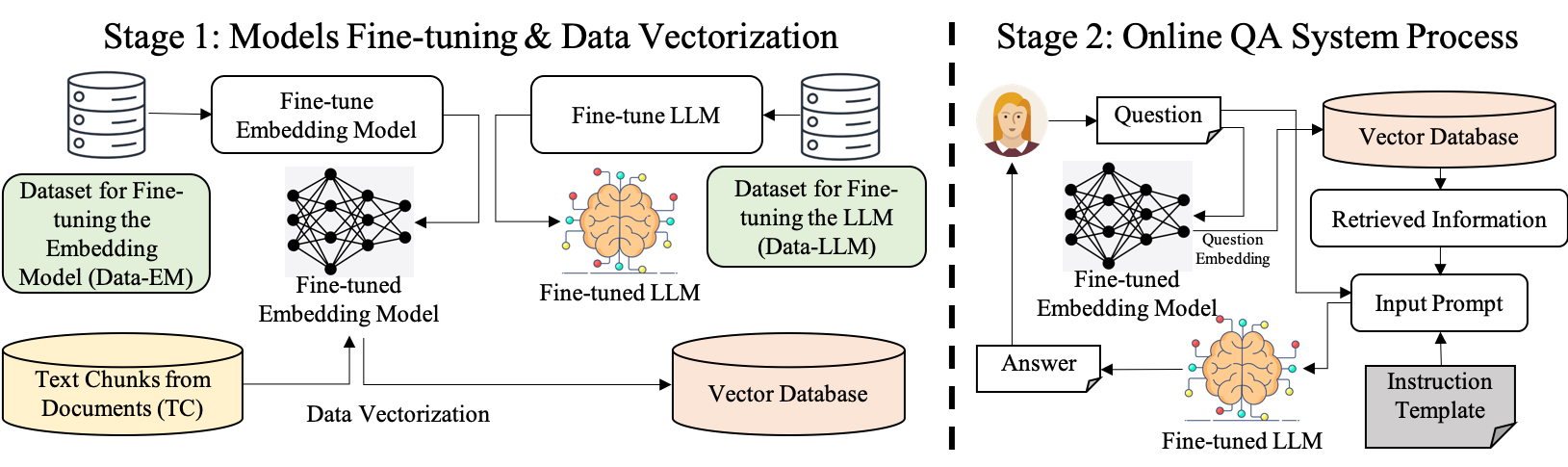}
  \vspace{-2ex}
  \caption{Overview of the proposed RAG4ITOps framework for IT operations and maintenance.}
  \label{framework}
  \vspace{-2ex}
\end{figure*}

\section{Related work}

\paragraph{IT Operations and Maintenance.} 
Traditionally, the quality of IT operations and maintenance varies because it highly depends on the IT operators' personal experience~\citep{notaro2020systematic}. To cultivate IT operators and meanwhile manage the ever-increasing IT-related information and knowledge well, QA systems are essential to improve efficiency across various application scenarios, developed by leveraging the development of NLP techniques~\citep{huang2023survey,elhoone2020cyber}. These systems aim to help IT operators quickly access useful information and develop troubleshooting plans~\citep{rijal2022aiops}. However, in practice, the IT operators may interact with the QA systems by several times to make a plan for difficult tasks like troubleshooting, because current QA systems are not intelligent enough to provide a comprehensive solution answer just within once interaction.

\paragraph{Large Language Models.} 
Recent LLMs have demonstrated significant advancements in open-domain QA tasks~\citep{brown2020language,openai2023gpt}. As to those closed-source models, like GPT-4, Claude and Gemini, they cannot answer domain-specific or even enterprise-exclusive questions well since they do not trained on any private documents. The other thing is that those open-source models, like LLaMA-3~\citep{touvron2023llama}, Qwen, and ChatGLM3~\citep{zeng2022glm, du2022glm}, can be directly fine-tuned on specific corpora and then provide QA services. However there are two major concerns. Firstly, LLMs often tend to generate hallucinated information~\citep{guo2023loglg}, which is unbearable in industrial area. Secondly, faced with the ever-increasing massive data, the QA systems based on LLMs have to be refine-tuned frequently, leading to a much high expense. Therefore, intuitively, RAG frameworks can remove the concerns and strengthen the LLMs-based QA systems. A recent effort to develop domain-specific LLMs, such as OWL~\citep{guo2023owl}, have shown promise. But it still struggles to be adaptive for grounded QA scenarios in real-world industrial IT operations.

\paragraph{Retrieval Augmented Generation.} To address the limitations of LLMs in factual issue and domain-specific applications, the RAG framework has emerged as a promising approach~\citep{gao2023retrieval}. RAG techniques aim to enhance the capabilities of LLMs by incorporating relevant external information into the input queries, thereby improving the accuracy and factuality of generated responses. In many domain-specific applications, RAG has proven highly effective for modeling domain-related semantics and improving the LLMs to output factual and satisfactory answers~\citep{gupta2024rag, wang2024unims, zhang2024raft}. Recent researches have further expanded RAG's potential, exploring the fine-tuning methods of pretrained LLMs specifically for RAG tasks~\citep{lin2023ra, zhang2024raft, wang2023instructretro, xu2023retrieval}. This paper also follows the idea of RAG, while we propose a more comprehensive and practical RAG framework specific for the domain of IT operations and maintenanc
\section{Methodology}

To facilitate the business process of data modeling and model fine-tuning of QA systems for IT operations and maintenance, we present the RAG4ITOps framework and introduce its details in this section. As shown in Figure~\ref{framework}, the framework includes two stages. One is for offline model fine-tuning and data vectorization, and the other is about the online QA system process based on RAG mechanism.

\subsection{Data Preprocessing}
\label{method:data}

Data preprocessing is particularly important for enterprise-level applications. Due to data privacy and data heterogeneity, a good data processing pipeline is essential to generate high-quality dataset for downstream model training. In terms of IT operations and maintenance, as instanced in Figure~\ref{task}, there are some characteristics of enterprise-exclusive terminologies, multi-source and multi-form documents and extremely long texts (e.g., texts about error log analysis and solution). Thus, we design a pipeline to preprocess the data.

\begin{figure}[t]
\centering
  \includegraphics[width=0.45\textwidth]{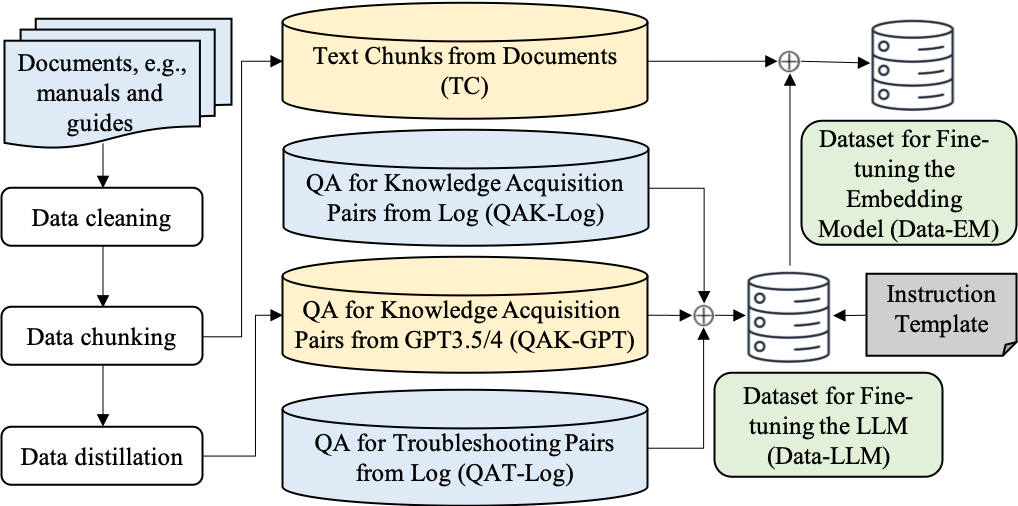}
  \vspace{-2ex}
  \caption{Data preprocessing method in our framework: datasets with a blue background originate from enterprise-exclusive corpora, those with a yellow background are post-preprocessed, and those with a green background are used for model fine-tuning.}
  \label{data}
  \vspace{-2ex}
\end{figure}

As shown in Figure~\ref{data}, the raw enterprise data colored with blue background are documents (e.g., manuals and guides), QA pairs for knowledge acquisition from log (QAK-Log), and QA pairs for troubleshooting from log (QAT-Log). Firstly, with the documents, the pipeline includes three main phases: data cleaning, data chunking and data distillation. Data cleaning is the process of removing unrelated tokens from documents. Data chunking splits long documents to shorter chunks (e.g., each chunk is less than 800 words), and data distillation generates more QA data with GPT-3.5/4.

Secondly, after data chunking, we obtain the dataset of text chunks from documents (TC). After data distillation, we get a dataset of QA for knowledge acquisition pairs from GPT-3.5/4 (QAK-GPT). By combining the QAK-Log, QAK-GPT and QAT-Log datasets, we create a dataset for fine-tuning the LLM (called Data-LLM) and a dataset for fine-tuning the embedding model (called Data-EM). Note that we design some instruction templates in advance and wrap the Data-LLM for training the ability of instruction compliance. After the data preprocessing pipeline, we obtained various datasets with their statistics summarized in Table~\ref{statistics}. All the details and data examples can be found in the Appendix section~\ref{app:dataset}.

\subsection{Instruction Template Design}
To effectively guide the LLM in generating appropriate responses for different QA tasks, we designed specific instruction templates. These templates serve to structure the input and provide task-specific context to the model. More detailed information on the specific prompts used can be found in the appendix section \ref{app:prompt}.

\subsection{Stage 1: Models Fine-tuning \& Data Vectorization}
\label{method:embed}

With the preprocessed text chunks and two datasets, the embedding model and LLM can be fine-tuned to better adapt for the enterprise-exclusive semantics and QA tasks. As shown in Figure~\ref{framework}, the Data-EM dataset is used to fine-tune the embedding model, while the Data-LLM dataset is used to fine-tune the LLM. Especially with the fine-tuned embedding model, the text chunks dataset can be vectorized as embeddings which are stored in the vector database for later online retrieval.

\subsubsection{Fine-tuning Embedding Model}
More technically, during fine-tuning the embedding model, we employ the Dense Passage Retrieval (DPR) framework~\citep{karpukhin-etal-2020-dense} as our base retrieval method. DPR uses embedding models to generate dense vector representations of both queries and passages, enabling efficient and accurate retrieval. Specifically, we begin with the pretrained embedding model, BGE-M3~\citep{bge_embedding}, known for its compact size and high performance on the MTEB benchmark~\citep{muennighoff2022mteb}. To further enhance its retrieval performance, we conducted contrastive learning with two kinds of negative sampling strategies, ensuring it effectively distinguishes between domain relevant and non-relevant passages.

\paragraph{Homogeneous In-Batch Negative Sampling (HIS).} To ensure the discriminative capability of the embeddings, a significant number of negative samples is necessary~\citep{qu2020rocketqa, wang2022text}. While in-batch negative sampling is a standard approach for introducing a substantial number of these samples, it comes with a drawback in our specific scenario: Negative samples from various tasks might not effectively distinguish semantic relationships within a particular context. To address this challenge, we structure each mini-batch to contain training data solely from identical tasks, thus maintaining homogeneity among the in-batch negatives and enhancing their contribution to the embeddings' discriminative ability. Our methodology incorporates both in-batch and hard negatives. Additionally, we utilize cross-device sharing~\citep{xiao2021matching} to increase the volume of negative samples available.

\paragraph{Auxiliary Hard Negative Sampling (AHNS).} Given the IT operations and maintenance dataset \( \mathcal{X} \), we aim to define an encoding function \( f : \mathcal{X} \to \mathbb{R}^d \) that assigns each document chunk or question \( x_i \in \mathcal{X} \) to a position in a \( d \)-dimensional embedding space. The goal is for the embeddings of related chunks and questions \( (x_i, x_i') \) to be proximate, and those of unrelated ones to be distant. For a random subset (batch) of \( N \) positive pairs \( \mathcal{X}_N = \{(\bar{x}_i, \tilde{x}_i)\}_{i=1}^N \), where \( \bar{x}_i, \tilde{x}_i \) represents a document chunk and its corresponding question. we define the contrastive loss function for the encoder \( f \) as follows:
\begin{equation}
\resizebox{0.85\hsize}{!}{
$\mathcal{L}_{x_i} = -\log \frac{\exp (s(\bar{x}_i, \tilde{x}_i) / \tau)}{\exp (s(\bar{x}_i, \tilde{x}_i) / \tau) + \sum_{\tilde{x}_j \in \mathcal{X}_N} \exp (s(\bar{x}_i, \tilde{x}_j) / \tau)},$
}
\end{equation}

where $s(x_i, x_j)=\frac{f(x_i)^\top f(x_j)}{\|f(x_i)\|\|f(x_j)\|}$ represents the inner product of the normalized latent representations of \(x_i \) and \( x_j \), and \( \tau \) is a temperature scaling hyperparameter. \( \tilde{x}_i \) is the positive sample associated with \( \bar{x}_i \) and all other instances \( \tilde{x}_j \neq \tilde{x}_i \in \mathcal{X}_N \) are considered negative samples. We aim to select high-quality, informative hard negative examples from this set. Typically, negative examples are chosen through random sampling~\citep{chen2020improved, chen2020simple}. 
Our approach employs the DPR framework with the initial embedding model to retrieve top-k relevant chunks for each positive sample. We then designate all remaining chunks, excluding the actual document chunks, as hard negative samples

\subsubsection{Fine-tuning LLM}
For fine-tuning the LLM of RAG4ITOps, we use a state-of-the-art LLM Qwen-14b-Base~\citep{bai2023qwen} as the backbone. Also we leverage two training methods to enhance the LLM's ability.

\begin{table*}
\centering
\resizebox{\linewidth}{!}{
\begin{tabular}{lcccc|ccc}
\hline
\multirow{2}{*}{Method} & \multirow{2}{*}{Supported Max Length}  & \multicolumn{3}{c}{QA for Knowledge Acquisition}  & \multicolumn{3}{c}{QA for Troubleshooting} \\
\cline{3-8}
                &                                                        & Acc@1 & Acc@5 & Acc@20  & Acc@1 & Acc@5 & Acc@20 \\
\hline
Text2Vec-base~\citep{text2vec}                                          &512   & 0.314 & 0.496 & 0.606 & 0.735 & 0.771 & 0.771  \\
M3E-base~\citep{Moka_Massive_Mixed_Embedding}                           &512   & 0.305 & 0.572 & 0.758 & 0.639 & 0.735 & 0.771  \\
GTE-large-zh~\citep{li2023towards}                                       &512   & 0.487 & 0.708 & 0.822 & 0.554 & 0.687 & 0.747  \\
BGE-large-zh-v1.5~\citep{bge_embedding}                                  &512   & 0.525 & 0.767 & 0.902 & 0.602 & 0.723 & 0.735  \\
jina-embeddings-v2-base-zh~\citep{mohr2024multi}                         &8192  & 0.369 & 0.674 & 0.847 & 0.566 & 0.747 & 0.783  \\
BGE-M3~\citep{bge-m3}                                                    &8192  & 0.610 & 0.881 & 0.958 & 0.651 & 0.759 & 0.783  \\
RAG4ITOps (Ours)                                                         &8192  & 0.661 & 0.919 & 0.979 & 0.759 & 0.795 & 0.795 \\
\hline
\end{tabular}
}
\vspace{-2ex}
\caption{Comparison of the fine-tuned embedding model with baselines. Acc@K represents top-K retrieval accuracy.}
\label{embedding-model}
\vspace{-2ex}
\end{table*} 

\paragraph{Continue Pre-Training} 
With the preprocessed domain-specific datasets, we aim to imbue the Qwen-14b-base model with specialized knowledge in IT operations and maintenance, enhancing its ability to understand and generate relevant content in this domain. The method is aligned with the standard approach~\citep{gururangan2020don}.

\paragraph{Retrieval Augmented Fine-Tuning Method} 
To enhance the LLM's ability to utilize retrieved information in IT operations and maintenance tasks, we implement a retrieval-augmented fine-tuning approach. Based on the Data-LLM dataset, we construct an extended training dataset (Data-LLM) \( D = \{(x^{(i)} \circ I^{(i)}, y^{(i)})\}_{i=1}^M \), where \( x^{(i)} \circ I^{(i)} \) represents an input query \( x^{(i)} \) accompanied by retrieved chunks \( I^{(i)} \), and \( y^{(i)} \) represents the output answer. 

For each example \((x^{(i)}, y^{(i)}) \in D\), we retrieve the top-k relevant text chunks \(I^{(i)} \subset C\) based on \(x^{(i)}\). We then create the fine-tuning instances by combining each retrieved chunk with the question using an instruction template (detailed in Appendix~\ref{app:prompt}).

The objective function of this supervised instruction tuning can be denoted as:
\begin{equation}
\resizebox{0.85\hsize}{!}{
$\mathcal{L}_m = -\frac{1}{N} \sum_{i=1}^{N} \mathbb{E}_{x, y, I \in D_i} \log P(y^{(i)} | I^{(i)} \circ x^{(i)}),$
}
\end{equation}

\noindent where \(P(y^{(i)} | I^{(i)} \circ x^{(i)})\) is the probability of generating the correct output \(y^{(i)}\) given the input \(x^{(i)}\) augmented with the retrieved chunks \(I^{(i)}\). This approach offers two key benefits: it adapts the LLM to utilize relevant and latest background knowledge, and it enables the LLM to generate factual answers.

\subsection{Stage 2: Online QA System Process}
\label{method:core}

At Stage 2, as shown in Figure~\ref{framework}, the IT operators can ask a question. Then the fine-tuned embedding model transforms the question into an embedding and the embedding is used to retrieve relevant contents from the vector database. We leverage FAISS~\citep{johnson2019billion}, a library for efficient similarity search, to identify the most relevant document chunks. With the retrieved information and question, they are wrapped by the instruction template to construct the input prompt for LLM. Finally, the LLM can answer the question by referring to all the contents in the input prompt. The whole process follows the prevailing RAG mechanism and achieves efficient response time.

\section{Experiment}


\subsection{Evaluation Dataset}
We collect a dataset called Data-Eval for evaluation. It comprises 319 questions created by domain experts, among which 236 questions for the knowledge acquisition task and 83 for the troubleshooting task. Each question is paired with relevant chunks from the enterprise-exclusive corpora, and all questions have labbeled answers.

\begin{table}
\setlength\tabcolsep{3pt}
\scriptsize
\centering
\begin{tabular}{ccccc|ccc}
\hline
\multirow{2}{*}{HIS} & \multirow{2}{*}{AHNS} & \multicolumn{3}{c}{QA for KA}  & \multicolumn{3}{c}{QA for TS} \\
\cline{3-8}
                              & & Acc@1 & Acc@5 & Acc@20  & Acc@1 & Acc@5 & Acc@20 \\
\hline
- & -                     & 0.661 & 0.895 & 0.970 & 0.711 & 0.771 & 0.790  \\
+ & -                      & 0.650 & 0.903 & 0.974 & 0.735 & 0.783 & 0.795  \\
- & +                      & 0.665 & 0.915 & 0.970 & 0.721 & 0.783 & 0.795 \\ 
+ & +                       & 0.661 & 0.919 & 0.979 & 0.759 & 0.795 & 0.795 \\
\hline
\end{tabular}
\vspace{-2ex}
\caption{Ablation study results for the fine-tuned embedding model in RAG4ITOps.}
\label{ab-embedding}
\vspace{-2ex}
\end{table} 

\begin{table}
\setlength\tabcolsep{2pt}
\scriptsize
\centering
\begin{tabular}{ccc|ccc}
\hline
\multicolumn{3}{c|}{QA for KA} & \multicolumn{3}{c}{QA for TS} \\
\hline
Chunks@1 & Chunks@5 & Chunks@20 & Chunks@1 & Chunks@5 & Chunks@20 \\
\hline
15.4 & 30.9 & 46.7 & 27.7 & 55.6 & 84.2 \\
\hline
\end{tabular}
\vspace{-2ex}
\caption{Response time(ms) for once retrieval.}
\label{inference-time}
\vspace{-5ex}
\end{table}

\begin{table*}
\centering
\small
\begin{tabular}{lcccc|cccc}
\hline
\multirow{2}{*}{Method}  & \multicolumn{4}{c}{QA for Knowledge Acquisition}  & \multicolumn{4}{c}{QA for Troubleshooting} \\
\cline{2-9}
& Score1 & Score2 & Score3 & Mean & Score1 & Score2 & Score3 & Mean \\
\hline
Chatglm3-6b~\citep{du2022glm}               & 5.19 & 5.28 & 5.20 & 5.22 & 4.01 & 4.06 & 4.26 & 4.11 \\
Qwen-7b-Chat~\citep{bai2023qwen}            & 5.89 & 5.84 & 5.80 & 5.84 & 5.21 & 5.37 & 5.22 & 5.27 \\
Llama3-8B-Instruct~\citep{touvron2023llama} & 5.32 & 5.23 & 5.40 & 5.32 & 5.61 & 5.68 & 5.62 & 5.64 \\
Qwen-14b-Chat~\citep{bai2023qwen}           & 6.57 & 6.58 & 6.63 & 6.59 & 5.99 & 6.07 & 6.10 & 6.05 \\
RAG4ITOps (Ours)                            & 6.92 & 7.01 & 6.70 & 6.88 & 6.72 & 6.65 & 6.68 & 6.68 \\
\hline
\end{tabular}
\vspace{-2ex}
\caption{Results of single-score mode evaluation on the fine-tuned LLM. Score1-3 mean that the GPT-4 are called for three times to evaluate each case.}
\label{single-score-mode}
\vspace{-2ex}
\end{table*} 

\begin{table}
\setlength\tabcolsep{2pt}
\centering
\scriptsize
\begin{tabular}{cccccc|cccc}
\hline
\multirow{2}{*}{CPT}  &\multirow{2}{*}{RAFT} & \multicolumn{4}{c}{QA for KA}  & \multicolumn{4}{c}{QA for TS} \\
\cline{3-10}
& & Score1 & Score2 & Score3 & Mean & Score1 & Score2 & Score3 & Mean \\
\hline
- & -  & 6.57 & 6.65 & 6.61 & 6.61 & 6.15 & 6.10 & 6.08 & 6.11 \\
+ & - & 6.62 & 6.61 & 6.68 & 6.64 & 6.14 & 6.15 & 6.10 & 6.13  \\
- & +  & 6.66 & 6.63 & 6.72 & 6.67 & 6.58 & 6.75 & 6.62 & 6.65 \\
+ & +   & 6.92 & 7.01 & 6.70 & 6.88 & 6.72 & 6.65 & 6.68 & 6.68 \\
\hline
\end{tabular}
\vspace{-2ex}
\caption{Ablation study results for the fine-tuned LLM.}
\label{ab-core}
\vspace{-4ex}
\end{table}

\subsection{Baselines and Metrics}
We consider the following popular text embedding models as the baselines for our embedding model evaluation: GTE-large-zh, BGE-M3, Text2Vec-base, M3E-base, jina-embeddings-v2-base-zh, and BGE-large-zh-v1.5. For the LLM evaluation, our method are compared with several state-of-the-art language models: Chatglm3-6b, Qwen-7b-Chat, Llama3-8B-Instruct, and Qwen-14b-Chat.

To evaluate the effectiveness of embedding model in the knowledge acquisition and troubleshooting tasks, we assessed performance using the top-k retrieval accuracy (Acc@K).
The formal definition of Acc@K can be defined as follows: 
\(R(q, C) \rightarrow \hat{C}\) takes as input question \(q\) and chunks \(C\) and returns a much smaller set \( \hat{C}\), where \(\hat{C} \subseteq C\) and \(|\hat{C}| = k \ll |C|\). Top-k retrieval accuracy is the fraction of questions for which \(\hat{C}\) contains a span that can answer the question. In our experiments, we separately present the results of log retrieval where the $k$ is set 1, 5 or 20.

To assess the performance of LLM, we employ two evaluation methods: single-score mode and pairwise-score mode~\citep{huang2024c, xu2023baize, guo2023owl, zheng2024judging}. In Single-score  mode, we first select the model to be tested and generate answers based on given questions and fixed reference chunks, using the BGE-M3 embedding model as default. We then utilize GPT-4~\citep{openai2023gpt} as a scoring model to evaluate the responses on a scale of 1 to 10, with higher scores indicating better quality. To ensure reliability, we run GPT-4 three times for each response, and report the average score in our results. In pairwise-score mode, both models generate answers to identical questions using the same reference chunks. A scoring model then assesses which model's responses are superior, assigning a win to the better performer and a loss to the other. If the performance is comparable, both models receive a tie. Detailed prompts and procedures for both evaluation modes are provided in the Appendix~\ref{app:prompt}.

\subsection{Evaluation Results for Embedding Model}
In Table~\ref{embedding-model}, our domain knowledge augmented embedding model demonstrates superior performance compared to baseline models across both tasks. 

Specifically for the QA for Knowledge Acquisition task (QA for KA), our full model with Homogeneous In-Batch Sampling (HIS) and Auxiliary Hard Negative Sampling (AHNS) achieves the highest Acc@5 and Acc@20 scores of 0.919 and 0.979 respectively, outperforming the BGE-M3 baseline by 4.3\% and 2.2\% on these metrics. In the QA for Troubleshooting task (QA for TS), our full model demonstrates the strongest performance, achieving the highest scores across all metrics: Acc@1 of 0.759, Acc@5 of 0.795, and Acc@20 of 0.795. These results represent improvements of 16.6\%, 4.7\%, and 1.5\% respectively over the BGE-M3 baseline.

Additionally, we also evaluated the inference time of our embedding model on an A100 80G GPU, as shown in Table~\ref{inference-time}, and the results demonstrate the efficiency of our method.

\subsection{Evaluation Results for LLM}
\begin{figure}
  \centering
  \scriptsize
  \includegraphics[width=0.48\textwidth, height=1\textheight, keepaspectratio]{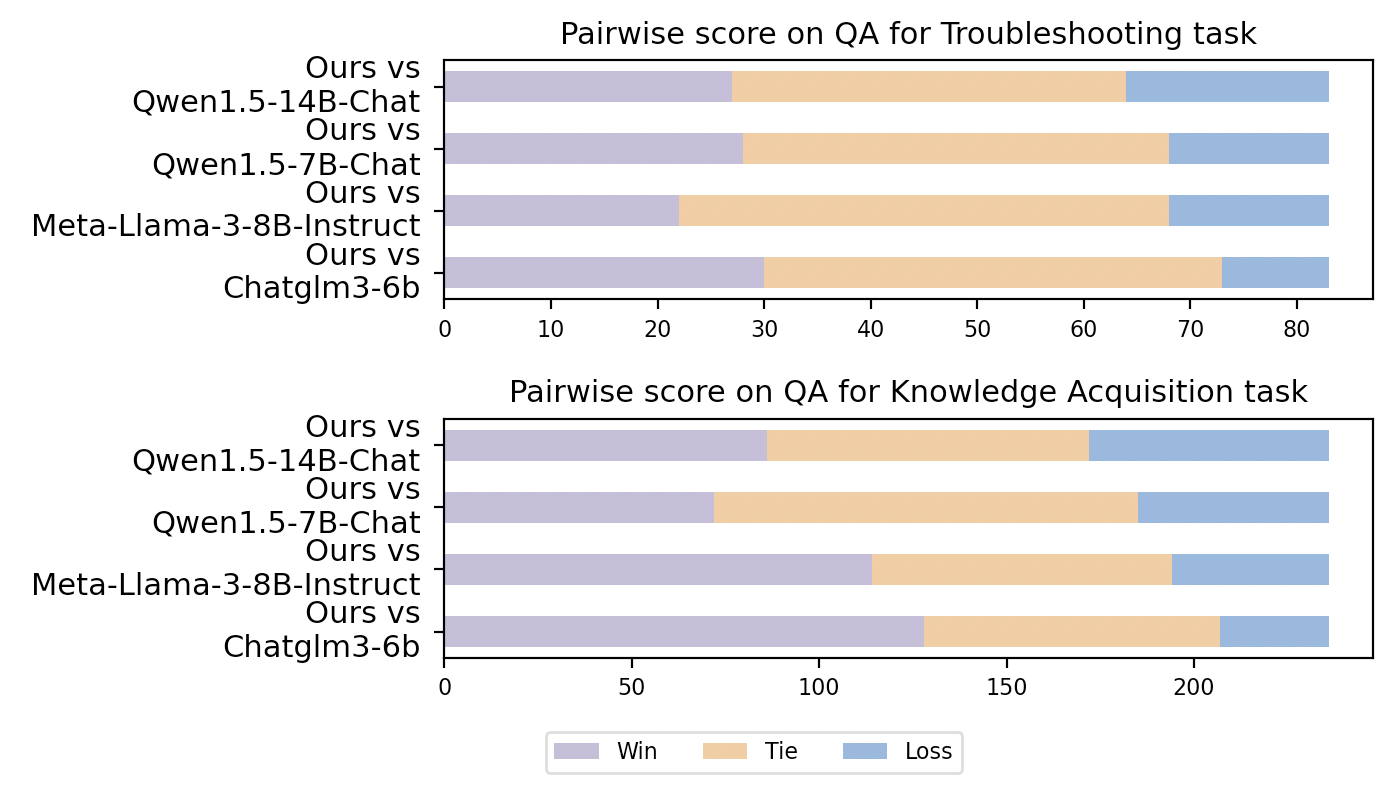}
  \vspace{-5ex}
  \caption{Pairwise comparison of our LLM against baselines in two tasks, evaluated by GPT-4.
}
  \label{pairwise_score}
\vspace{-5ex}
\end{figure}

For single-score mode, we compared our proposed model against several baseline models, including Chatglm3-6b, Qwen-7b-Chat, Llama3-8B-Instruct, and Qwen-14b-Chat. Table \ref{single-score-mode} shows that our model with Continue Pre-Training (CPT) and Retrieval Augmented Fine-Tuning Method (RAFT) achieves the highest mean scores in both QA for Troubleshooting (6.68) and QA for Knowledge Acquisition (6.88) tasks. These scores represent improvements of 0.63 and 0.29 points respectively over the Qwen-14b-Chat baseline. As for the pairwise scores (see Figure \ref{pairwise_score}), our model outperforms all baselines in both tasks.

\subsection{Ablation study}
For the embedding model, we evaluated the impact of HIS and AHNS. Results in Table~\ref{ab-embedding} show that both techniques contribute to performance gains, with their combination yielding the best results across all metrics in both tasks.

For the LLM, we conducted an ablation study to examine the importance of CPT and RAFT. In Table~\ref{single-score-mode}, the baseline model scored 6.05 for QA for Troubleshooting and 6.59 for QA for Knowledge Acquisition. In Table~\ref{ab-core}, Supervised Fine-Tuning without chunks (w/o CPT w/o RAFT) showed improvements over the baseline. RAFT alone (w/o CPT w/ RAFT) further improved scores to 6.65 and 6.67, outperforming standard Supervised Fine-Tuning and demonstrating its effectiveness in enhancing model performance. Our full model, incorporating both CPT and RAFT, achieved the highest scores of 6.68 for Troubleshooting and 6.88 for Knowledge Acquisition. This represents noticeable improvements of 0.57 points (9.3\%) for Troubleshooting and 0.27 points (4.1\%) for Knowledge Acquisition compared to the model (w/o CPT w/o RAFT), highlighting the complementary benefits of our proposed techniques

\section{Conclusion}
In this paper, we introduce RAG4ITOps, a comprehensive framework for QA systems tailored for IT operations and maintenance. Initially, we developed a dataset construction pipeline, incorporating data cleaning, chunking, and distillation of enterprise-exclusive corpora. Additionally, we fine-tuned an embedding model and enhanced its retrieval performance using Homogeneous In-Batch Negative Sampling and Auxiliary Hard Negative Sampling strategies. Furthermore, we leveraged and fine-tuned a LLM enhancing its capabilities for domain-specific QA tasks with Continue Pre-Training and Retrieval Augmented Fine-Tuning. We evaluated our framework through a series of experiments, designed to assess its performance on distinct QA tasks with different difficulties, demonstrating the effectiveness of our approach in the domain of IT operations and maintenance.

\section*{Acknowledgement}

This work is supported by 2024 Ningbo ``Innovation Yongjiang 2035'' Key Technology Breakthrough Programme (No. 2024Z119). 
We thank all the anonymous reviewers for their insightful and constructive comments.

\bibliography{acl_latex}

\begin{thebibliography}{46}
\providecommand{\natexlab}[1]{#1}

\bibitem[{Achiam et~al.(2023)Achiam, Adler, Agarwal, Ahmad, Akkaya, Aleman,
  Almeida, Altenschmidt, Altman, Anadkat et~al.}]{openai2023gpt}
Josh Achiam, Steven Adler, Sandhini Agarwal, Lama Ahmad, Ilge Akkaya,
  Florencia~Leoni Aleman, Diogo Almeida, Janko Altenschmidt, Sam Altman,
  Shyamal Anadkat, et~al. 2023.
\newblock \href {https://arxiv.org/abs/2303.08774} {Gpt-4 technical report}.
\newblock \emph{Preprint}, arXiv:2303.08774.

\bibitem[{Bai et~al.(2023)Bai, Bai, Chu, Cui, Dang, Deng, Fan, Ge, Han, Huang
  et~al.}]{bai2023qwen}
Jinze Bai, Shuai Bai, Yunfei Chu, Zeyu Cui, Kai Dang, Xiaodong Deng, Yang Fan,
  Wenbin Ge, Yu~Han, Fei Huang, et~al. 2023.
\newblock \href {https://arxiv.org/abs/2309.16609} {Qwen technical report}.
\newblock \emph{Preprint}, arXiv:2309.16609.

\bibitem[{Brown et~al.(2020)Brown, Mann, Ryder, Subbiah, Kaplan, Dhariwal,
  Neelakantan, Shyam, Sastry, Askell et~al.}]{brown2020language}
Tom Brown, Benjamin Mann, Nick Ryder, Melanie Subbiah, Jared~D Kaplan, Prafulla
  Dhariwal, Arvind Neelakantan, Pranav Shyam, Girish Sastry, Amanda Askell,
  et~al. 2020.
\newblock Language models are few-shot learners.
\newblock \emph{Advances in neural information processing systems},
  33:1877--1901.

\bibitem[{Chen et~al.(2024)Chen, Xiao, Zhang, Luo, Lian, and Liu}]{bge-m3}
Jianlv Chen, Shitao Xiao, Peitian Zhang, Kun Luo, Defu Lian, and Zheng Liu.
  2024.
\newblock \href {https://arxiv.org/abs/2402.03216} {Bge m3-embedding:
  Multi-lingual, multi-functionality, multi-granularity text embeddings through
  self-knowledge distillation}.
\newblock \emph{Preprint}, arXiv:2402.03216.

\bibitem[{Chen et~al.(2020{\natexlab{a}})Chen, Kornblith, Norouzi, and
  Hinton}]{chen2020simple}
Ting Chen, Simon Kornblith, Mohammad Norouzi, and Geoffrey Hinton.
  2020{\natexlab{a}}.
\newblock A simple framework for contrastive learning of visual
  representations.
\newblock In \emph{International conference on machine learning}, pages
  1597--1607.

\bibitem[{Chen et~al.(2020{\natexlab{b}})Chen, Fan, Girshick, and
  He}]{chen2020improved}
Xinlei Chen, Haoqi Fan, Ross Girshick, and Kaiming He. 2020{\natexlab{b}}.
\newblock \href {https://arxiv.org/abs/2003.04297} {Improved baselines with
  momentum contrastive learning}.
\newblock \emph{Preprint}, arXiv:2003.04297.

\bibitem[{Chowdhery et~al.(2023)Chowdhery, Narang, Devlin, Bosma, Mishra,
  Roberts, Barham, Chung, Sutton, Gehrmann et~al.}]{chowdhery2023palm}
Aakanksha Chowdhery, Sharan Narang, Jacob Devlin, Maarten Bosma, Gaurav Mishra,
  Adam Roberts, Paul Barham, Hyung~Won Chung, Charles Sutton, Sebastian
  Gehrmann, et~al. 2023.
\newblock Palm: Scaling language modeling with pathways.
\newblock \emph{Journal of Machine Learning Research}, 24(240):1--113.

\bibitem[{Du et~al.(2017)Du, Li, Zheng, and Srikumar}]{du2017deeplog}
Min Du, Feifei Li, Guineng Zheng, and Vivek Srikumar. 2017.
\newblock Deeplog: Anomaly detection and diagnosis from system logs through
  deep learning.
\newblock In \emph{Proceedings of the 2017 ACM SIGSAC conference on computer
  and communications security}, pages 1285--1298.

\bibitem[{Du et~al.(2022)Du, Qian, Liu, Ding, Qiu, Yang, and Tang}]{du2022glm}
Zhengxiao Du, Yujie Qian, Xiao Liu, Ming Ding, Jiezhong Qiu, Zhilin Yang, and
  Jie Tang. 2022.
\newblock Glm: General language model pretraining with autoregressive blank
  infilling.
\newblock In \emph{Proceedings of the 60th Annual Meeting of the Association
  for Computational Linguistics (Volume 1: Long Papers)}, pages 320--335.

\bibitem[{Elhoone et~al.(2020)Elhoone, Zhang, Anwar, and
  Desai}]{elhoone2020cyber}
Hietam Elhoone, Tianyang Zhang, Mohd Anwar, and Salil Desai. 2020.
\newblock Cyber-based design for additive manufacturing using artificial neural
  networks for industry 4.0.
\newblock \emph{International Journal of Production Research},
  58(9):2841--2861.

\bibitem[{Galup et~al.(2009)Galup, Dattero, Quan, and
  Conger}]{galup2009overview}
Stuart~D Galup, Ronald Dattero, Jim~J Quan, and Sue Conger. 2009.
\newblock An overview of it service management.
\newblock \emph{Communications of the ACM}, 52(5):124--127.

\bibitem[{Gao et~al.(2021)Gao, Yao, and Chen}]{gao2021simcse}
Tianyu Gao, Xingcheng Yao, and Danqi Chen. 2021.
\newblock \href {https://arxiv.org/abs/2104.08821} {Simcse: Simple contrastive
  learning of sentence embeddings}.
\newblock \emph{Preprint}, arXiv:2104.08821.

\bibitem[{Gao et~al.(2023)Gao, Xiong, Gao, Jia, Pan, Bi, Dai, Sun, and
  Wang}]{gao2023retrieval}
Yunfan Gao, Yun Xiong, Xinyu Gao, Kangxiang Jia, Jinliu Pan, Yuxi Bi, Yi~Dai,
  Jiawei Sun, and Haofen Wang. 2023.
\newblock \href {https://arxiv.org/abs/2312.10997} {Retrieval-augmented
  generation for large language models: A survey}.
\newblock \emph{Preprint}, arXiv:2312.10997.

\bibitem[{Guo et~al.(2023)Guo, Guo, Yang, Liu, Li, Zheng, Zheng, Hou, and
  Zhang}]{guo2023loglg}
Hongcheng Guo, Yuhui Guo, Jian Yang, Jiaheng Liu, Zhoujun Li, Tieqiao Zheng,
  Liangfan Zheng, Weichao Hou, and Bo~Zhang. 2023.
\newblock Loglg: Weakly supervised log anomaly detection via log-event graph
  construction.
\newblock In \emph{International Conference on Database Systems for Advanced
  Applications}, pages 490--501.

\bibitem[{Guo et~al.(2024)Guo, Yang, Liu, Yang, Chai, Bai, Peng, Hu, Chen,
  Zhang, xu~Shi, Zheng, liangfan zheng, Zhang, Xu, and Li}]{guo2023owl}
Hongcheng Guo, Jian Yang, Jiaheng Liu, Liqun Yang, Linzheng Chai, Jiaqi Bai,
  Junran Peng, Xiaorong Hu, Chao Chen, Dongfeng Zhang, xu~Shi, Tieqiao Zheng,
  liangfan zheng, Bo~Zhang, Ke~Xu, and Zhoujun Li. 2024.
\newblock {OWL}: A large language model for {IT} operations.
\newblock In \emph{The Twelfth International Conference on Learning
  Representations}.

\bibitem[{Gupta et~al.(2024)Gupta, Shirgaonkar, Balaguer, Silva, Holstein, Li,
  Marsman, Nunes, Rouzbahman, Sharp et~al.}]{gupta2024rag}
Aman Gupta, Anup Shirgaonkar, Angels de~Luis Balaguer, Bruno Silva, Daniel
  Holstein, Dawei Li, Jennifer Marsman, Leonardo~O Nunes, Mahsa Rouzbahman,
  Morris Sharp, et~al. 2024.
\newblock \href {https://arxiv.org/abs/2401.08406} {Rag vs fine-tuning:
  Pipelines, tradeoffs, and a case study on agriculture}.
\newblock \emph{Preprint}, arXiv:2401.08406.

\bibitem[{Gururangan et~al.(2020)Gururangan, Marasovi{\'c}, Swayamdipta, Lo,
  Beltagy, Downey, and Smith}]{gururangan2020don}
Suchin Gururangan, Ana Marasovi{\'c}, Swabha Swayamdipta, Kyle Lo, Iz~Beltagy,
  Doug Downey, and Noah~A Smith. 2020.
\newblock \href {https://arxiv.org/abs/2004.10964} {Don't stop pretraining:
  Adapt language models to domains and tasks}.
\newblock \emph{Preprint}, arXiv:2004.10964.

\bibitem[{Hu et~al.(2021)Hu, Shen, Wallis, Allen-Zhu, Li, Wang, Wang, and
  Chen}]{hu2021lora}
Edward~J Hu, Yelong Shen, Phillip Wallis, Zeyuan Allen-Zhu, Yuanzhi Li, Shean
  Wang, Lu~Wang, and Weizhu Chen. 2021.
\newblock \href {https://arxiv.org/abs/2106.09685} {Lora: Low-rank adaptation
  of large language models}.
\newblock \emph{Preprint}, arXiv:2106.09685.

\bibitem[{Huang et~al.(2023{\natexlab{a}})Huang, Yu, Ma, Zhong, Feng, Wang,
  Chen, Peng, Feng, Qin et~al.}]{huang2023survey}
Lei Huang, Weijiang Yu, Weitao Ma, Weihong Zhong, Zhangyin Feng, Haotian Wang,
  Qianglong Chen, Weihua Peng, Xiaocheng Feng, Bing Qin, et~al.
  2023{\natexlab{a}}.
\newblock \href {https://arxiv.org/abs/2311.05232} {A survey on hallucination
  in large language models: Principles, taxonomy, challenges, and open
  questions}.
\newblock \emph{Preprint}, arXiv:2311.05232.

\bibitem[{Huang et~al.(2023{\natexlab{b}})Huang, Liu, Fung, Qi, Yang, and
  Luan}]{huang2023logqa}
Shaohan Huang, Yi~Liu, Carol Fung, Jiaxing Qi, Hailong Yang, and Zhongzhi Luan.
  2023{\natexlab{b}}.
\newblock \href {https://arxiv.org/abs/2303.11715} {Logqa: Question answering
  in unstructured logs}.
\newblock \emph{Preprint}, arXiv:2303.11715.

\bibitem[{Huang et~al.(2024)Huang, Bai, Zhu, Zhang, Zhang, Su, Liu, Lv, Zhang,
  Fu et~al.}]{huang2024c}
Yuzhen Huang, Yuzhuo Bai, Zhihao Zhu, Junlei Zhang, Jinghan Zhang, Tangjun Su,
  Junteng Liu, Chuancheng Lv, Yikai Zhang, Yao Fu, et~al. 2024.
\newblock C-eval: A multi-level multi-discipline chinese evaluation suite for
  foundation models.
\newblock \emph{Advances in Neural Information Processing Systems}, 36.

\bibitem[{J{\"a}ntti and Cater-Steel(2017)}]{jantti2017proactive}
Marko J{\"a}ntti and Aileen Cater-Steel. 2017.
\newblock Proactive management of it operations to improve it services.
\newblock \emph{JISTEM-Journal of Information Systems and Technology
  Management}, 14(2):191--218.

\bibitem[{Johnson et~al.(2019)Johnson, Douze, and
  J{\'e}gou}]{johnson2019billion}
Jeff Johnson, Matthijs Douze, and Herv{\'e} J{\'e}gou. 2019.
\newblock Billion-scale similarity search with gpus.
\newblock \emph{IEEE Transactions on Big Data}, 7(3):535--547.

\bibitem[{Karpukhin et~al.(2020)Karpukhin, Oguz, Min, Lewis, Wu, Edunov, Chen,
  and Yih}]{karpukhin-etal-2020-dense}
Vladimir Karpukhin, Barlas Oguz, Sewon Min, Patrick Lewis, Ledell Wu, Sergey
  Edunov, Danqi Chen, and Wen-tau Yih. 2020.
\newblock Dense passage retrieval for open-domain question answering.
\newblock In \emph{Proceedings of the 2020 Conference on Empirical Methods in
  Natural Language Processing (EMNLP)}, pages 6769--6781.

\bibitem[{Li et~al.(2023)Li, Zhang, Zhang, Long, Xie, and
  Zhang}]{li2023towards}
Zehan Li, Xin Zhang, Yanzhao Zhang, Dingkun Long, Pengjun Xie, and Meishan
  Zhang. 2023.
\newblock \href {https://arxiv.org/abs/2308.03281} {Towards general text
  embeddings with multi-stage contrastive learning}.
\newblock \emph{Preprint}, arXiv:2308.03281.

\bibitem[{Lin et~al.(2023)Lin, Chen, Chen, Shi, Lomeli, James, Rodriguez, Kahn,
  Szilvasy, Lewis et~al.}]{lin2023ra}
Xi~Victoria Lin, Xilun Chen, Mingda Chen, Weijia Shi, Maria Lomeli, Rich James,
  Pedro Rodriguez, Jacob Kahn, Gergely Szilvasy, Mike Lewis, et~al. 2023.
\newblock \href {https://arxiv.org/abs/2310.01352} {Ra-dit: Retrieval-augmented
  dual instruction tuning}.
\newblock \emph{Preprint}, arXiv:2310.01352.

\bibitem[{Liu et~al.(2023)Liu, Tao, Meng, Wang, Ma, Zhao, Chen, Yang, Jiang,
  and Chen}]{liu2023logprompt}
Yilun Liu, Shimin Tao, Weibin Meng, Jingyu Wang, Wenbing Ma, Yanqing Zhao,
  Yuhang Chen, Hao Yang, Yanfei Jiang, and Xun Chen. 2023.
\newblock \href {https://arxiv.org/abs/2308.07610} {Logprompt: Prompt
  engineering towards zero-shot and interpretable log analysis}.
\newblock \emph{Preprint}, arXiv:2308.07610.

\bibitem[{Mohr et~al.(2024)Mohr, Krimmel, Sturua, Akram, Koukounas,
  G{\"u}nther, Mastrapas, Ravishankar, Mart{\'\i}nez, Wang
  et~al.}]{mohr2024multi}
Isabelle Mohr, Markus Krimmel, Saba Sturua, Mohammad~Kalim Akram, Andreas
  Koukounas, Michael G{\"u}nther, Georgios Mastrapas, Vinit Ravishankar,
  Joan~Fontanals Mart{\'\i}nez, Feng Wang, et~al. 2024.
\newblock \href {https://arxiv.org/abs/2402.17016} {Multi-task contrastive
  learning for 8192-token bilingual text embeddings}.
\newblock \emph{Preprint}, arXiv:2402.17016.

\bibitem[{Muennighoff et~al.(2022)Muennighoff, Tazi, Magne, and
  Reimers}]{muennighoff2022mteb}
Niklas Muennighoff, Nouamane Tazi, Lo{\"\i}c Magne, and Nils Reimers. 2022.
\newblock \href {https://arxiv.org/abs/2210.07316} {Mteb: Massive text
  embedding benchmark}.
\newblock \emph{Preprint}, arXiv:2210.07316.

\bibitem[{Notaro et~al.(2020)Notaro, Cardoso, and
  Gerndt}]{notaro2020systematic}
Paolo Notaro, Jorge Cardoso, and Michael Gerndt. 2020.
\newblock A systematic mapping study in aiops.
\newblock In \emph{International Conference on Service-Oriented Computing},
  pages 110--123.

\bibitem[{Qu et~al.(2020)Qu, Ding, Liu, Liu, Ren, Zhao, Dong, Wu, and
  Wang}]{qu2020rocketqa}
Yingqi Qu, Yuchen Ding, Jing Liu, Kai Liu, Ruiyang Ren, Wayne~Xin Zhao, Daxiang
  Dong, Hua Wu, and Haifeng Wang. 2020.
\newblock \href {https://arxiv.org/abs/2010.08191} {Rocketqa: An optimized
  training approach to dense passage retrieval for open-domain question
  answering}.
\newblock \emph{Preprint}, arXiv:2010.08191.

\bibitem[{Rijal et~al.(2022)Rijal, Colomo-Palacios, and
  S{\'a}nchez-Gord{\'o}n}]{rijal2022aiops}
Laxmi Rijal, Ricardo Colomo-Palacios, and Mary S{\'a}nchez-Gord{\'o}n. 2022.
\newblock Aiops: A multivocal literature review.
\newblock \emph{Artificial Intelligence for Cloud and Edge Computing}, pages
  31--50.

\bibitem[{Touvron et~al.(2023)Touvron, Martin, Stone, Albert, Almahairi,
  Babaei, Bashlykov, Batra, Bhargava, Bhosale et~al.}]{touvron2023llama}
Hugo Touvron, Louis Martin, Kevin Stone, Peter Albert, Amjad Almahairi, Yasmine
  Babaei, Nikolay Bashlykov, Soumya Batra, Prajjwal Bhargava, Shruti Bhosale,
  et~al. 2023.
\newblock \href {https://arxiv.org/abs/2307.09288} {Llama 2: Open foundation
  and fine-tuned chat models}.
\newblock \emph{Preprint}, arXiv:2307.09288.

\bibitem[{Wang et~al.(2023{\natexlab{a}})Wang, Ping, McAfee, Xu, Li, Shoeybi,
  and Catanzaro}]{wang2023instructretro}
Boxin Wang, Wei Ping, Lawrence McAfee, Peng Xu, Bo~Li, Mohammad Shoeybi, and
  Bryan Catanzaro. 2023{\natexlab{a}}.
\newblock \href {https://arxiv.org/abs/2310.07713} {Instructretro: Instruction
  tuning post retrieval-augmented pretraining}.
\newblock \emph{Preprint}, arXiv:2310.07713.

\bibitem[{Wang et~al.(2024)Wang, Huang, Deng, Wang, Wang, Wang, Mi, Pan, and
  Wong}]{wang2024unims}
Hongru Wang, Wenyu Huang, Yang Deng, Rui Wang, Zezhong Wang, Yufei Wang, Fei
  Mi, Jeff~Z Pan, and Kam-Fai Wong. 2024.
\newblock \href {https://arxiv.org/abs/2401.13256} {Unims-rag: A unified
  multi-source retrieval-augmented generation for personalized dialogue
  systems}.
\newblock \emph{Preprint}, arXiv:2401.13256.

\bibitem[{Wang et~al.(2022)Wang, Yang, Huang, Jiao, Yang, Jiang, Majumder, and
  Wei}]{wang2022text}
Liang Wang, Nan Yang, Xiaolong Huang, Binxing Jiao, Linjun Yang, Daxin Jiang,
  Rangan Majumder, and Furu Wei. 2022.
\newblock \href {https://arxiv.org/abs/2212.03533} {Text embeddings by
  weakly-supervised contrastive pre-training}.
\newblock \emph{Preprint}, arXiv:2212.03533.

\bibitem[{Wang et~al.(2023{\natexlab{b}})Wang, Sun, and
  He}]{Moka_Massive_Mixed_Embedding}
Y~Wang, Q~Sun, and S~He. 2023{\natexlab{b}}.
\newblock M3e: Moka massive mixed embedding model.
\newblock \url{https://github. com/wangyuxinwhy/uniem}.

\bibitem[{Xiao et~al.(2021)Xiao, Liu, Shao, Lian, and Xie}]{xiao2021matching}
Shitao Xiao, Zheng Liu, Yingxia Shao, Defu Lian, and Xing Xie. 2021.
\newblock \href {https://arxiv.org/abs/2104.07858} {Matching-oriented product
  quantization for ad-hoc retrieval}.
\newblock \emph{Preprint}, arXiv:2104.07858.

\bibitem[{Xiao et~al.(2023)Xiao, Liu, Zhang, and Muennighoff}]{bge_embedding}
Shitao Xiao, Zheng Liu, Peitian Zhang, and Niklas Muennighoff. 2023.
\newblock \href {https://arxiv.org/abs/2309.07597} {C-pack: Packaged resources
  to advance general chinese embedding}.
\newblock \emph{Preprint}, arXiv:2309.07597.

\bibitem[{Xu et~al.(2023{\natexlab{a}})Xu, Guo, Duan, and
  McAuley}]{xu2023baize}
Canwen Xu, Daya Guo, Nan Duan, and Julian McAuley. 2023{\natexlab{a}}.
\newblock \href {https://arxiv.org/abs/2304.01196} {Baize: An open-source chat
  model with parameter-efficient tuning on self-chat data}.
\newblock \emph{Preprint}, arXiv:2304.01196.

\bibitem[{Xu(2023)}]{text2vec}
Ming Xu. 2023.
\newblock Text2vec: Text to vector toolkit.
\newblock \url{https://github.com/shibing624/text2vec}.

\bibitem[{Xu et~al.(2023{\natexlab{b}})Xu, Ping, Wu, McAfee, Zhu, Liu,
  Subramanian, Bakhturina, Shoeybi, and Catanzaro}]{xu2023retrieval}
Peng Xu, Wei Ping, Xianchao Wu, Lawrence McAfee, Chen Zhu, Zihan Liu, Sandeep
  Subramanian, Evelina Bakhturina, Mohammad Shoeybi, and Bryan Catanzaro.
  2023{\natexlab{b}}.
\newblock \href {https://arxiv.org/abs/2310.03025} {Retrieval meets long
  context large language models}.
\newblock \emph{Preprint}, arXiv:2310.03025.

\bibitem[{Zeng et~al.(2022)Zeng, Liu, Du, Wang, Lai, Ding, Yang, Xu, Zheng, Xia
  et~al.}]{zeng2022glm}
Aohan Zeng, Xiao Liu, Zhengxiao Du, Zihan Wang, Hanyu Lai, Ming Ding, Zhuoyi
  Yang, Yifan Xu, Wendi Zheng, Xiao Xia, et~al. 2022.
\newblock \href {https://arxiv.org/abs/2210.02414} {Glm-130b: An open bilingual
  pre-trained model}.
\newblock \emph{Preprint}, arXiv:2210.02414.

\bibitem[{Zhang et~al.(2023)Zhang, Xiao, Liu, Dou, and Nie}]{zhang2023retrieve}
Peitian Zhang, Shitao Xiao, Zheng Liu, Zhicheng Dou, and Jian-Yun Nie. 2023.
\newblock \href {https://arxiv.org/abs/2310.07554} {Retrieve anything to
  augment large language models}.
\newblock \emph{Preprint}, arXiv:2310.07554.

\bibitem[{Zhang et~al.(2024)Zhang, Patil, Jain, Shen, Zaharia, Stoica, and
  Gonzalez}]{zhang2024raft}
Tianjun Zhang, Shishir~G Patil, Naman Jain, Sheng Shen, Matei Zaharia, Ion
  Stoica, and Joseph~E Gonzalez. 2024.
\newblock \href {https://arxiv.org/abs/2403.10131} {Raft: Adapting language
  model to domain specific rag}.
\newblock \emph{Preprint}, arXiv:2403.10131.

\bibitem[{Zheng et~al.(2024)Zheng, Chiang, Sheng, Zhuang, Wu, Zhuang, Lin, Li,
  Li, Xing et~al.}]{zheng2024judging}
Lianmin Zheng, Wei-Lin Chiang, Ying Sheng, Siyuan Zhuang, Zhanghao Wu, Yonghao
  Zhuang, Zi~Lin, Zhuohan Li, Dacheng Li, Eric Xing, et~al. 2024.
\newblock Judging llm-as-a-judge with mt-bench and chatbot arena.
\newblock \emph{Advances in Neural Information Processing Systems}, 36.

\end{thebibliography}

\appendix

\section{Appendix}
\label{sec:appendix}

\begin{table*}
\centering
\small
\begin{tabular}{|c|l|c|c|}
\hline
Dataset Name & Dataset Description & Indicator & Number \\
\hline
\multirow{2}{*}{TC} & \multirow{2}{*}{Text Chunks} & Chunks \# &  3,824    \\
\cline{3-4}
&& Avg. token \# & 529 \\
\hline
\multirow{3}{*}{QAK-Log} & \multirow{3}{*}{QA pairs for knowledge acquisition from log} & Sample \# & 1,468  \\
\cline{3-4}
&& Avg. question token \# & 16  \\
\cline{3-4}
&& Avg. answer token \# & 56 \\
\hline
\multirow{3}{*}{QAK-GPT} & \multirow{3}{*}{QA pairs for knowledge acquisition from GPT-3.5/4} & Sample \# & 16,973  \\
\cline{3-4}
&&  Avg. question token \# &  15  \\
\cline{3-4}
&& Avg. answer token \# &  66 \\
\hline
\multirow{3}{*}{QAT-Log} & \multirow{3}{*}{QA pairs for troubleshooting from log} & Sample \# &  47,471   \\
\cline{3-4}
&&  Avg. question token \# &  235 \\
\cline{3-4}
&&  Avg. answer token \# &  370 \\
\hline
\multirow{1}{*}{Data-EM} & Dataset for fine-tuning the embedding model & Sample \# &  65,912 \\
\hline
\multirow{1}{*}{Data-Pretrain} & Dataset for Pretrain the LLM & token \# &  1,604,448 \\
\hline
\multirow{3}{*}{Data-LLM} & \multirow{3}{*}{Dataset for fine-tuning the LLM} & Sample \# & 65,912 \\
\cline{3-4}
&&   Avg. question token \# & 1233 \\
\cline{3-4}
&&   Avg. answer token \# & 186 \\
\hline
\multirow{3}{*}{Data-Eval} & \multirow{3}{*}{Dataset for evaluation} & Sample \# & 319 \\
\cline{3-4}
&&   Avg. question token \# & 53 \\
\cline{3-4}
&&   Avg. answer token \# & 133 \\
\hline
\end{tabular}
\caption{Statistics of the datasets used in RAG4ITOps.}
\label{statistics}
\end{table*}

\subsection{Experimental Settings}
When fine-tuning the embedding model, the learning rate we set is $10^{-6}$, a batch size of 1024. We use Adam as the optimization algorithm with $\beta1$ = 0.9, $\beta2$ = 0.99. We also implemented the homogeneous in-batch sampling strategy, where all samples in the same batch come from the same task, and utilized negatives cross-device to enhance the diversity of negative samples. The model was trained for 1 epoch using 8*A100 80G GPUs.

For Continue Pre-Training the LLM, we set the learning rate to $2 \times 10^{-5}$, with a weight decay of 0.1, and global batch size of 128. The sequence length is set at 2048. We use the Adam optimization algorithm with $\beta_1 = 0.9$ and $\beta_2 = 0.99$. The training epoch is 3. For Retrieval Augmented Fine-Tuning, the learning rate is increased to $5 \times 10^{-5}$, maintaining the same weight decay of 0.1, the global batch size is 512. The sequence length remains 2048. Adam is again used as the optimization algorithm, with the same $\beta$ values. The training duration for this phase is 1 epoch. We conduct full parameter training for Continue Pre-Training using 8*A100 80G GPUs and LoRA~\citep{hu2021lora} fine-tuning for Retrieval Augmented Fine-Tuning using 8*A100 80G GPUs.

\subsection{Dataset Construction}
\label{app:dataset}
High-quality datasets are essential for effective Large Language Model(LLM) implementation, often more crucial than model architecture updates. With improved data collection and processing techniques, we can perform Continue Pre-Training and Retrieval Augmented Fine-Tuning (RAFT) on the model more effectively and achieve better Retrieval-Augmented Generation (RAG) performance. 
We designed a sophisticated dataset construction pipeline including phases such as collection, chunking, distillation, and combination. This pipeline is capable of extracting features from each type of data and provides robust support for the LLM to meet the specific requirements of the IT operations and maintenance group.

\minisection{IT Operations Data} This data was provided by the IT operations group and contains documents and QA pairs. The documents include Word files with internal knowledge such as tool descriptions, operation examples, system configurations, and scripts. These documents contain text, images, and tables. As we currently focus on language modeling, we only extracted texts and tables using the python-docx. 

\minisection{Maintenance Data} The maintenance group provided $47k$ pairs of error logs and corresponding analyses. The error logs contain detailed descriptions of errors, functions, and related platforms. The analyses are human-labeled and include error scenarios, problem localization, and solutions. Specifically, problem localization contains function names, function descriptions, error reasons, priorities, and impacts. 

\subsubsection{Data Processing}

\minisection{General Processing} We convert all information into text format to make documents easy to handle. By using python-docx, we fully extract all tables from Word files and convert them to plain text based on LaTeX standards. Each row is joined by a line break, and each column is joined by a vertical line. This approach enables the model to recognize all information within tables. Furthermore, we standardize texts by removing noisy tokens and converting illegal tokens to their normal forms. We use these processed texts from documents to form the pre-training dataset.

\minisection{Chunking Techniques} As documents often contain very long and complex structures, we split each document into several chunks. Chunking techniques are essential in our task. Complete and reasonable chunks can provide meaningful context to enhance performance in data distillation and data retrieval. Since most of the current documents are in a fixed format, we designed a targeted chunking method for these documents to achieve better results than general splitting methods. Moreover, we also designed a general chunking method for new incoming documents to do online training.

At the beginning of chunking, we first remove noisy content using heuristic methods. As each document contains a menu with clear signs, we explore the scope of menus and remove them all. Additionally, due to the presence of technical documents, we remove noisy sentences and tables containing words like "Script Maintainer" or "Version Number" which is only for human understanding and technical requirements, so we prevent the model from learning them to increase training and retrieval efficiency.

The targeted chunking method primarily focuses on maintaining the logical integrity of each sentence. Unlike setting a fixed length for each chunk, this method preserves the complete meaning and logic of contents as much as possible, especially for tables. It helps the LLM gain a comprehensive understanding of contents and avoid hallucinations due to forced sentence segmentation. As we extract all information from documents in Word file format, each line has its type, including content, style, and font. The style represents whether it is a title or normal text and the level of the title. Since each title signifies an individual block, we can separate the content based on title levels.

We start by splitting each content into blocks by 'Heading 1', the largest title. For each block, we count the number of tokens using the same tokenizer. If the number is less than 800, we consider this block as a whole and do not split it further. Conversely, we continue splitting the block by 'Heading 2', and so on. After recursion, if the number of tokens in a block exceeds 800 but cannot be split further, we resort to using the general method introduced below to split the sentence. By setting 800 as the threshold, we can include enough complete contexts in the RAG result.

Moreover, in our experiments, we find that the short sentence is ineffective for model understanding and affect data distillation performance. Therefore, we combine contents with fewer than 20 tokens into nearby blocks. We also separate titles and contents, underlining them using a template like "Title: <title> Content: <content>". This approach, similar to human reading patterns, allows the retrieval model to work effectively and easily find accurate results.

The general method is a default version that splits the document into blocks of nearly fixed length without considering its format. Generally, we split the document to ensure each sentence has fewer than 800 tokens and includes overlap between sentences. Furthermore, we make the ending token of each sentence a typical stop word such as a line break, dot, or comma, to ensure the sentence has complete meaning.

In our experiments, these two methods produce very reasonable chunks for most cases, as confirmed by human labeling. Using the methods described above, we collected $3k$ chunks from the documents and build the dataset TC.

\subsubsection{Data Distillation}
In addition to using RAG to enhance the LLM's understanding of documents, we also implement data distillation to generate a number of real-world cases and provide additional guidance to the model.

For chunks extracted from documents, we collect QA pairs from each chunk by calling the APIs of GPT-3.5 and GPT-4. These QA pairs simulate real questions and expected answers based on the documents. In the training phase, we combine the distilled questions and corresponding RAG contexts as input, and use the expected answers as output to maintain a data format similar to real cases.

To optimize costs, we typically call the GPT-3.5 API to generate instructions from the text of chunks $i$ times, where $i = round(2 + \frac{n-1000}{500})$ and $n$ is the number of characters. This dynamic generation method allows us to capture more information for long and complex sentences. The prompt used for data distillation is provided in the Appendix~\ref{app:prompt}.

If the response from GPT-3.5 has an incorrect format, we resort to calling the GPT-4 API to obtain a more accurate result. Through this data distillation process, we ultimately create the dataset (QAK-GPT) with $1.6k$ instructions, each containing questions, answers, and corresponding raw contents as context.

For the QA pairs from the raw data, we aim to use them for RAFT while preventing their questions from appearing in the context. To achieve this, we rewrite each question to form a RAFT dataset. The prompt used for this rewriting process is provided in the Appendix~\ref{app:prompt}.

In this way, we obtain a RAFT (QAK-Log) with different questions but the same answers. During the training phase, we can provide these questions and the raw QA pairs as input, expecting the model to learn to generate the correct answer as output.

\subsubsection{Data Combination}
As described above, for Continue Pre-Training, we directly use the texts extracted from documents (Data-Pretrain). For Retrieval Augmented Generation(RAG), we combine the text chunks from documents (TC), QA pairs for knowledge acquisition (QAK-Log and QAK-GPT), and QA pairs for troubleshooting (QAT-Log) to create the RAG dataset. We also combine the QAK-GPT, QAK-Log, and QAT-Log to form the final RAFT dataset (Data-LLM), containing $65k$ rows of data.

\subsection{Instruction Templates and Prompts}
\label{app:prompt}
This section presents the detailed prompts used in our question-answering system for IT operations and maintenance. Table \ref{prompt_table_1} and \ref{prompt_table_2} presents two key prompts used in our evaluation process: the Pairwise-Score Mode Prompt and the Single-score Mode Prompt.  
Table \ref{prompt_table_3} presents additional prompts used in our data preparation pipeline. The first prompt is designed for rewriting sentences while preserving their meaning, which is useful for data augmentation and diversity. The second prompt is used in our data distillation process. The third and forth prompts are instruction template.

\subsection{Case study}
\label{app:case}
To provide insight into real-world applications of our RAG4ITOps framework, we present two representative cases: one illustrating a QA scenario for troubleshooting as shown in Table~\ref{usercase-error}, and another demonstrating a QA scenario for knowledge acquisition in IT operations and maintenance as shown in Table~\ref{usercase-qa}.

\begin{table*}
\caption{Evaluation prompts for pairwise-score and single-score modes for QA Knowledge Acquisition Task}
\centering
\begin{tabular}{p{0.9\textwidth}}
\hline
\textbf{Pairwise-Score Mode Prompt:} \\
Please act as an impartial evaluator and assess the quality of answers provided by two AI assistants to a user's question. Your evaluation should consider the correctness and helpfulness of the answers. You will be given a reference answer, Assistant A's answer, and Assistant B's answer. Your task is to determine which assistant's answer is better.

Evaluation steps:

1. Compare both assistants' answers to the reference answer.\\
2. Identify and correct any errors in the assistants' answers.\\
3. Avoid any positional bias, ensuring that the order of the answers does not influence your decision.\\
4. Do not let the length of the answers affect your assessment.\\
5. Please answer based on facts, expressing the required information for the question.\\
6. Do not favor certain assistant names. Be as objective as possible.

After providing your explanation, please output your final verdict in the following JSON format:

```json \\
\{ \\
  "verdict": "Can only be A or B or Tie", \\
  "explanation": "Your explanation" \\
\} \\
```\\
\hline
\textbf{Single-score Mode Prompt:} \\
Please act as an impartial evaluator and assess the quality of an answer provided by an AI assistant to a user's question. We will provide a question, a corresponding reference answer, and the assistant's answer. Your evaluation should consider the correctness of the answer.

Evaluation steps:

1. Please compare the assistant's answer to the reference answer.\\
2. Identify and correct any errors.\\
3. Evaluate as objectively as possible, paying attention to factual errors in the assistant's answer that are not present in the reference answer.\\
4. If the assistant does not address the content of the reference answer, it will be scored as 0.\\

After providing your explanation, you must rate the answer on a scale of 1 to 10 using the following JSON format:

```json\\
\{\\
  "rating": "1 to 10",\\
  "explanation": "Your explanation"\\
\}\\
```\\
\end{tabular}
\label{prompt_table_1}
\end{table*}

\begin{table*}
\caption{Evaluation prompts for pairwise-score and single-score modes for QA Troubleshooting Task}
\centering
\begin{tabular}{p{0.9\textwidth}}
\hline
\textbf{Pairwise-Score Mode Prompt:} \\
Please act as an impartial evaluator and assess the quality of answers provided by two AI assistants to a user's question. Your evaluation should consider the correctness and helpfulness of the answers. You will be given a reference answer, Assistant A's answer, and Assistant B's answer. Your task is to determine which assistant's answer is better.

Evaluation steps:

1. Compare both assistants' answers to the reference answer.\\
2. Identify and correct any errors in the assistants' answers.\\
3. Avoid any positional bias, ensuring that the order of the answers does not influence your decision.\\
4. Do not let the length of the answers affect your assessment.\\
5. Please answer based on facts, expressing the required information for the question.\\
6. Do not favor certain assistant names. Be as objective as possible.\\
7. The reference answer includes 7 fields, each field is worth 1 point, with the solution field worth 4 points. Please strictly compare the answers of both assistants for each field and analyze them. Based on the field scores, determine which assistant's answer is better, or if it's a tie\\

After providing your explanation, please output your final verdict in the following JSON format:

```json\\
\{\\
  "verdict": "Can only be A or B or Tie",\\
  "explanation": "Your explanation"\\
\}\\
```\\
\hline
\textbf{Single-score Mode Prompt:} \\
Please act as an impartial evaluator and assess the quality of an answer provided by an AI assistant to a user's question. We will provide a question, a corresponding reference answer, and the assistant's answer. Your evaluation should consider the correctness of the answer.

Evaluation steps:

1. Please compare the assistant's answer to the reference answer.\\
2. Identify and correct any errors.\\
3. Evaluate as objectively as possible, paying attention to factual errors in the assistant's answer that are not present in the reference answer.\\
4. If the assistant does not address the content of the reference answer, it will be scored as 0.\\
5. The reference answer includes 7 fields, each field is worth 1 point, with the solution field worth 4 points. For each field in the assistant's answer, please strictly score according to the field score. If the answer is inaccurate or incorrect for a field, no points should be awarded for that field.\\

After providing your explanation, you must rate the answer on a scale of 1 to 10 using the following JSON format:

```json\\
\{\\
  "rating": "1 to 10",\\
  "explanation": "Your explanation"\\
\}\\
```\\
\end{tabular}
\label{prompt_table_2}
\end{table*}

\begin{table*}
\caption{Data preparation prompts for sentence rewriting and data distillation}
\centering
\begin{tabular}{p{0.9\textwidth}}
\hline
\textbf{The prompt used for rewriting process:} \\
Assume you are the IT operation team member. Please rewrite the following sentence without changing its meaning.

Content: $<$content$>$ \\
\hline
\textbf{The prompt used for data distillation:} \\
Assume you are the IT operation team member and you have some questions to inquire. Assume the following document can answer your question. What questions and corresponding answers can you post?

Please post as many knowledge based questions as possible.\\
Do not post the question without an answer.\\
Answer should be complete and must be got from the document.\\
Question with very long answer is allowed.\\
If you cannot find any question or cannot provide answer, please respond $<$unk$>$.\\
Use $<$sep$>$ to connect each QA.
Content: $<$content$>$\\
\hline
\textbf{The instruction template for Knowledge Acquisition task:} \\
Assume you are a customer service representative, and you have received a question from a user or the operations team:\\
Content: $<$content$>$\\
Please answer the user's question concisely and professionally based on the following known information:\\
Content: $<$content$>$\\

\hline
\textbf{The instruction template for QA Troubleshooting Task:} \\
Please conduct a root cause analysis of the sudden AIOPS event based on the error log below. The analysis should include: 1. Scenario, 2. Problem localization (including service, method name, function, keywords, event type, event level, impact scope), 3. Solution (including personnel involved and resolution plan).
Content: $<$content$>$\\
Below is a historical case: \\
Content: $<$content$>$\\
\end{tabular}
\label{prompt_table_3}
\end{table*}

\begin{table*}
\caption{Example of a QA for troubleshooting case in IT operations and maintenance}
\centering
\begin{tabular}{p{0.9\textwidth}}
\hline
\textbf{User questions:} \\
Please help analyze and resolve the following error: GetCustomBiz query for custom topology failed. The error is: Querying business: *'s custom topology node id: [] failed, context canceled, project\_space=* namespace.name=*. \\
\textbf{Core model input:} \\
Please conduct a root cause analysis of the sudden AIOPS event based on the error log below. The analysis should include: 1. Scenario, 2. Problem localization (including service, method name, function, keywords, event type, event level, impact scope), 3. Solution (including personnel involved and resolution plan).

GetCustomBiz query for custom topology failed. The error is: Querying business: *'s custom topology node id: [] failed, context canceled, project\_space=* namespace.name=*. \\

Below is a historical case: \\
Segment 0: Error log case and solution 1 \\
Segment 1: Error log case and solution 2 \\
Segment 2: Error log case and solution 3 \\
Segment 3: Error log case and solution 4 \\
Segment 4: Error log case and solution 5 \\

\textbf{System response:} \\
Scenario: * log error \\
Project: * \\
Service: * \\
Method name: * \\
Function description: Set custom business NodeName for clusters and modules \\
Key information: GetCustomBiz query for custom topology failed. The error is querying custom topology node failed, topology node id: [] \\

Event type: Query * timeout caused query failure \\
Event level: * \\
Impact scope: Unable to normally obtain custom business NodeName, which may affect cluster and other interface queries \\

Solution:\\
Based on the error information, the query for custom topology node failed because the topology node id is empty. Possible solutions are as follows:\\

1. Check if the topology node id parameter is correctly passed. Ensure that the passed topology node id is not empty and is in the expected format. \\
2. Check if the topology node data exists. Confirm that the data corresponding to the topology node id exists in the system and has not been deleted or modified. \\
3. Check the logic of querying the custom topology. Ensure that the code logic for querying the custom topology is correct and there are no missed or erroneous handling of the topology node id. \\
4. Check the status of * services. If querying the custom topology requires other services, such as *, ensure these services are running normally and can be accessed properly. \\
5. Check logs and error information. View more detailed error logs and error information to further locate the problem. It may be necessary to increase log recording or adjust the log level to obtain more detailed information. \\

\end{tabular}
\label{usercase-error}
\end{table*}

\begin{table*}
\caption{Example of knowledge acquisition QA scenario in IT operations and maintenance}
\centering
\begin{tabular}{p{0.9\textwidth}}
\hline
\textbf{User questions:} \\
In the fault handling process for localized service anomalies, what are the specific steps for collecting exception information? \\
\textbf{Core model input:} \\
Assume you are a customer service representative, and you have received a question from a user or the operations team:\\
In the fault handling process for localized service anomalies, what are the specific steps for collecting exception information?\\
Please answer the user's question concisely and professionally based on the following known information:\\
Segment 0: Internal customer document Segment 1 \\
Segment 1: Internal customer document Segment 2 \\
Segment 2: Internal customer document Segment 3 \\
Segment 3: Internal customer document Segment 4 \\
Segment 4: Internal customer document Segment 5 \\
\textbf{System response:} \\
Specific steps for collecting exception information include:\\
Organize effective information based on customer complaints/internal alerts: The dimensions are not limited to the following information; the more comprehensive the information, the better.\\
Attempt to analyze/reproduce the problem: Based on the information provided by the customer, manually designate nodes for testing. Simultaneously, use the internal platform Grafana to investigate quality data across multiple dimensions such as bandwidth, status codes, and response times by combining domain name, node, and time information. \\

\end{tabular}
\label{usercase-qa}
\end{table*}

\end{document}